\documentclass[letterpaper]{article} % DO NOT CHANGE THIS
\usepackage{aaai25}  % DO NOT CHANGE THIS
\usepackage{times}  % DO NOT CHANGE THIS
\usepackage{helvet}  % DO NOT CHANGE THIS
\usepackage{courier}  % DO NOT CHANGE THIS
\usepackage[hyphens]{url}  % DO NOT CHANGE THIS
\usepackage{graphicx} % DO NOT CHANGE THIS
\def\UrlFont{\rm}  % DO NOT CHANGE THIS
\usepackage{natbib}  % DO NOT CHANGE THIS AND DO NOT ADD ANY OPTIONS TO IT
\usepackage{caption} % DO NOT CHANGE THIS AND DO NOT ADD ANY OPTIONS TO IT
\usepackage{algorithmic}
\usepackage{multirow}
\usepackage{booktabs}
\usepackage{xcolor}
\usepackage{amssymb} 
\usepackage{xspace}
\usepackage{amsmath} 
\usepackage{floatrow}
\newcommand{\ourmodel}{{HIP}\xspace}
\usepackage[linesnumbered,ruled,vlined]{algorithm2e}
\usepackage[switch]{lineno}

\usepackage[T1]{fontenc}
\usepackage{textcomp}
\usepackage{newfloat}
\usepackage{listings}
\DeclareCaptionStyle{ruled}{labelfont=normalfont,labelsep=colon,strut=off} % DO NOT CHANGE THIS
\floatstyle{ruled}
\newfloat{listing}{tb}{lst}{}
\floatname{listing}{Listing}
\nocopyright

\title{HIP: Hierarchical Point Modeling and Pre-training for Visual \\
Information Extraction}

\author{
~~Rujiao Long\footnotemark[1],\quad Pengfei Wang\footnotemark[1],\quad Zhibo Yang\footnotemark[2],\quad Cong Yao
}
\affiliations{
\textbf{Alibaba Group}
}

\usepackage{bibentry}
\begin{document}

% \title{HIP: Hierarchical Point Modeling and Pre-training for Visual \\
% Information Extraction} 
\maketitle

\renewcommand{\thefootnote}{\fnsymbol{footnote}}
\footnotetext[1]{Equal Contribution.} 
\footnotetext[2]{Correspondence Author.}

\begin{abstract}
End-to-end visual information extraction (VIE) aims at integrating the hierarchical subtasks of VIE, including text spotting, word grouping, and entity labeling, into a unified framework. Dealing with the gaps among the three subtasks plays a pivotal role in designing an effective VIE model. OCR-dependent methods heavily rely on offline OCR engines and inevitably suffer from OCR errors, while OCR-free methods, particularly those employing a black-box model, might produce outputs that lack interpretability or contain hallucinated content. Inspired by CenterNet, DeepSolo, and ESP,  we propose \ourmodel, which models entities as \textbf{HI}erarchical \textbf{P}oints to better conform to the hierarchical nature of the end-to-end VIE task.
Specifically, such \textit{hierarchical points} can be flexibly encoded and subsequently decoded into desired text transcripts, centers of various regions, and categories of entities. Furthermore, we devise corresponding hierarchical pre-training strategies, categorized as image reconstruction, layout learning, and language enhancement, to reinforce the cross-modality representation of the hierarchical encoders. Quantitative experiments on public benchmarks demonstrate that \ourmodel outperforms previous state-of-the-art methods, while qualitative results show its excellent interpretability.   
\end{abstract}

\vspace{-5mm}
\section{Introduction}
\label{sec:intro}

Visual information extraction (VIE) from scanned documents and captured photos has gained increasing attention because of its wide range of applications in various domains. Typical applications usually involve receipts
of shopping, business cards, product manuals, and bills of entry \cite{yang2023modeling}. While VIE greatly frees efforts from manually processing a great number of documents, it is accompanied by stringent requirements and a low tolerance for errors.

The latest research trend in VIE focuses on closing the gap between stages in the pipeline: namely text spotting \cite{huang2022swintextspotter, ye2023deepsolo}, word grouping \cite{yu2023structextv2}, and entity labeling \cite{zhang2020trie, xu2020layoutlm, yang2023modeling}. These three tasks are mutually interdependent: incorrect text spotting results in erroneous entity values, while inaccurate word grouping leads to imprecise entity boundaries. 
OCR-dependent methods, such as LayoutLM~\cite{xu2020layoutlm} and its variants \cite{xu2020layoutlmv2,huang2022layoutlmv3,xu2021layoutxlm, gu2022xylayoutlm}, rely heavily on the OCR engines and inevitably suffer from
OCR errors. OCR-free methods~\cite{Kim2021DonutDU}, which usually employ a black-box model, might generate outputs that lack interpretability or may contain hallucinated content.

\begin{figure}[t]
  \centering  \includegraphics[width=1\textwidth]{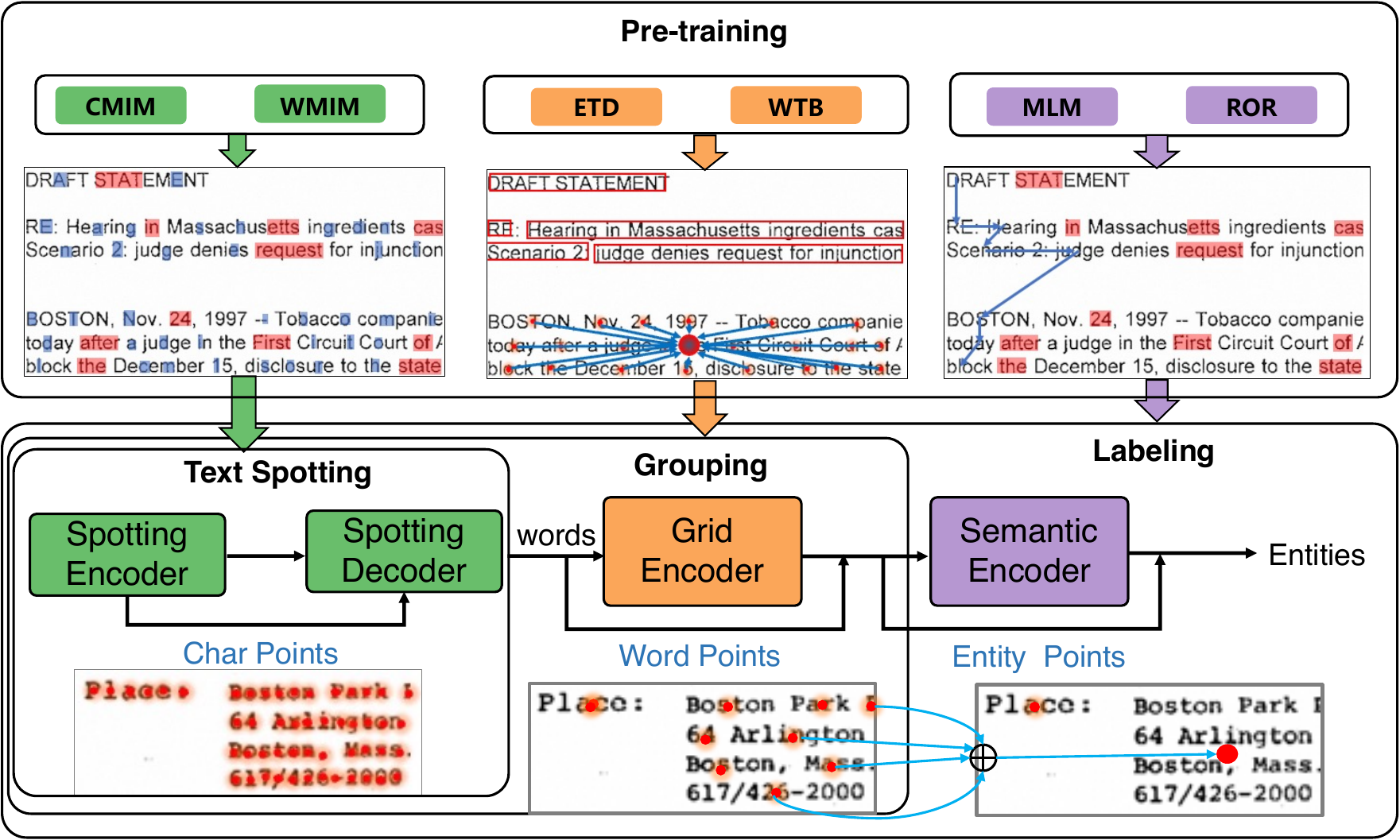}
  \vspace{-6mm}  
\caption{Illustration of the proposed HIP framework and the idea of hierarchical points. VIE is decomposed into three successive tasks: word spotting, word grouping, and entity labeling, each of which is accompanied by two pre-training strategies to learn visual, geometric, and semantic clues.} 
  \label{fig:fig1}

\end{figure}
To enable early mitigation of OCR errors, joint learning methods incorporate OCR as an integral part, and they allow for synergistic optimization from text detection to information extraction. TRIE \cite{zhang2020trie} and VIES \cite{wang2021towards} both designed networks for training multiple tasks simultaneously. They both explicitly classified multi-modal information to get final entities without any general visual-language pre-training, which may lead to weak linguistic representation. StrucTexTv2 \cite{yu2023structextv2} proposes a stronger representation for document understanding, but it requires additional decoding for OCR and VIE.

In this paper, we propose \ourmodel, a joint learning framework for visual information extraction by hierarchically modeling entities as character-level, word-level, and entity-level points. As shown in Fig.\ref{fig:fig1}, we first represent each character sequence as ordered points. After the decoding stage of the text spotter, we obtain the requisite contents and locations at the word level. Then, we take the central point of each word as a word-level representation. By calculating the average of all the central points within an entity, we derive an entity-level point that serves as the representation of the entity. Compared to other text representations
~\cite{liao2020mask, Wang2020AETL}, the point representation serves as a more flexible intermediary~\cite{duan2019centernet,ye2023deepsolo}, whether it is used for text spotting or word grouping.

To enhance the cross-modality representation of the hierarchical encoders, we propose corresponding hierarchical pre-training strategies: (1) Regarding the spotting task, char-level and word-level mask image modeling tasks (corresponding to CMIM and WMIM) are introduced to reconstruct character and word regions of images, respectively. CMIM and WMIM effectively enhance the model's ability to accurately recognize individual characters and grasp the semantic context of the text. (2) For the grouping task,  bottom-up (Word to Block, WTB) and top-down (Entity Detection, ETD) tasks are devised to encourage the model to fully learn layout information. (3) In the entity labeling stage, mask language modeling (MLM) and reading order reasoning (ROR) tasks are employed to strengthen the model's linguistic capabilities. 

Although the previous work ESP~\cite{yang2023modeling} represents entities as points, there is only one entity granularity in ESP, and it is unable to perform token-level recognition and more flexible entity boundary distinction. In contrast, our method yields better recognition results and a more flexible entity granularity.

Ultimately, we achieved state-of-the-art~(SOTA) performance on the FUNSD dataset, surpassing StrucTextv2~\cite{yu2023structextv2} by over 5.2\%, which is a new record. On CORD and SROIE, our method also obtains SOTA performance in terms of end-to-end F-score. The major contributions of \ourmodel are summarized as follows:

% \vspace{-3mm}
\begin{itemize}
    \item[-] {We propose \ourmodel, a joint learning framework for VIE. \ourmodel models entities as hierarchical points to better fit the hierarchical nature of the VIE task.}
%    \item[-] {We introduce a stage-wise encoder that leverages hierarchical points as an intermediary, which are compatible with various type of decoding.}
    \item[-] {We devise hierarchical pre-training strategies, categorized as image reconstruction, layout learning, and language enhancement, which facilitate the model to better learn complex visual, geometric and semantic clues.}
    \item[-] {Experiments on typical VIE benchmarks demonstrate the effectiveness and interpretability of the proposed \ourmodel.}     
\end{itemize}

% \vspace{-6mm}
\section{Related Work}
% \vspace{-1mm}
\subsection{Text Spotting} 
As text detection \cite{shi2017detecting,liao2020real} and text recognition \cite{shi2016end,shi2018aster,fang2021read} technologies gradually mature, the task of end-to-end text detection and recognition, known as text spotting \cite{li2017towards}, is increasingly becoming a focus of research. Most pioneering works \cite{he2018end, liao2020mask, liu2018fots, liu2020abcnet, wang2021pan++}  in text spotting follow a pipeline that first detects and then recognizes text, whereby the features of the Region-of-Interest (ROI) are passed to the recognizer. Segmentation-based approaches \cite{wang2021pgnet, xing2019convolutional} attempt to implement detection and recognition on different branches based on a shared backbone. Transformer-based text spotting methods \cite{huang2022swintextspotter, kittenplon2022towards, peng2022spts} have subsequently emerged. Specifically, \cite{ye2023deepsolo} proposes a novel query form based on explicit points sampled from the Bezier center curve representation of text instance lines, which can efficiently encode the position, shape, and semantics of text.
DeepSolo \cite{ye2023deepsolo} models text lines in a manner consistent with our approach, and we both utilize point representations.
% \vspace{-3mm}
\subsection{Pre-training Tasks}
% Mask modeling first demonstrates its powerful capabilities in BERT \cite{devlin2018bert}. BERT
Multimodal pre-training has flourished over the past few years, in which knowledge is learned from large-scale unlabeled data.
The BERT \cite{devlin2018bert} introduced the Mask Language Modeling (MLM) pre-training task, where the model is asked to infer the masked tokens. %Training on a massive dataset, the MLM task substantially enhanced the representation power of the language model, and become an indispensable task in various language pre-training models. 
Inspired by the MLM task,  the MAE \cite{he2022masked} and LORE++ \cite{long2025lore++} transfer to visual representation by predicting masked image patches. MAE has also been incorporated into text recognition task \cite{guan2023self, jiang2023revisiting}. In LayoutLMv2~\cite{xu2020layoutlmv2} and  LayoutXLM ~\cite{xu2021layoutxlm}, a Text-Image Alignment task is adopted to capture the fine-grained alignment relationship between text and image. By randomly selecting texts and masking the corresponding image regions, the model predicts whether the corresponding image region has been masked. LayoutLMv3 \cite{huang2022layoutlmv3} introduces a Word-Patch Alignment (WPA) objective to learn a fine-grained alignment between text words and image patches. The WPA objective predicts whether the corresponding image patch of a text word is masked. By reconstructing masked words in images with an additional ROI Alignment module, StrucTextv2 \cite{yu2023structextv2} learns a better document understanding.

% \vspace{-3mm}
\subsection{Entity Extraction}
Entity extraction aims at automatically acquiring key information from visually-rich documents. The language-based methods \cite{li2021structext, gu2022xylayoutlm} serializes the offline OCR of the entire document in a top-to-bottom, left-to-right order to a long sequence, and the entity extraction task is converted into BIO tagging for each token in the sequence. 
% However, patching and serializing images weakens their perception of visual positions. as a result, many studies have been conducted to improve position embeddings \cite{gu2022xylayoutlm}.
Other methods~\cite{lee2021rope, lee2022formnet, lee2023formnetv2} attempt to model entity relationships based on graph convolutional networks, and they are quite similar to the language-based methods. As discussed in ESP~\cite{yang2023modeling}, these methods primarily evaluate entity tagging accuracy under given ground truth, hence achieving high metrics. However, their end-to-end performances are significantly lower.
%Moreover, since the transformer itself can construct relationships, there are relatively fewer works based on GCN . 
The end-to-end encoder-decoder approaches \cite{dhouib2023docparser,kim2022ocr} employ learnable queries and features to perform cross-attention, and the decoder produces the category, content, and other information. Compared to language-based methods, end-to-end approaches consider the complete process of entity extraction, and the difficulty is significantly increased. Our work focuses on the more challenging end-to-end VIE task.

% \vspace{-3mm}
\section{Methodology}
\label{sec:method}
In this section, we will first introduce the overall framework and the hierarchical pre-training tasks. 
\begin{figure*}[tb]
  \centering
  \includegraphics[height=5cm]{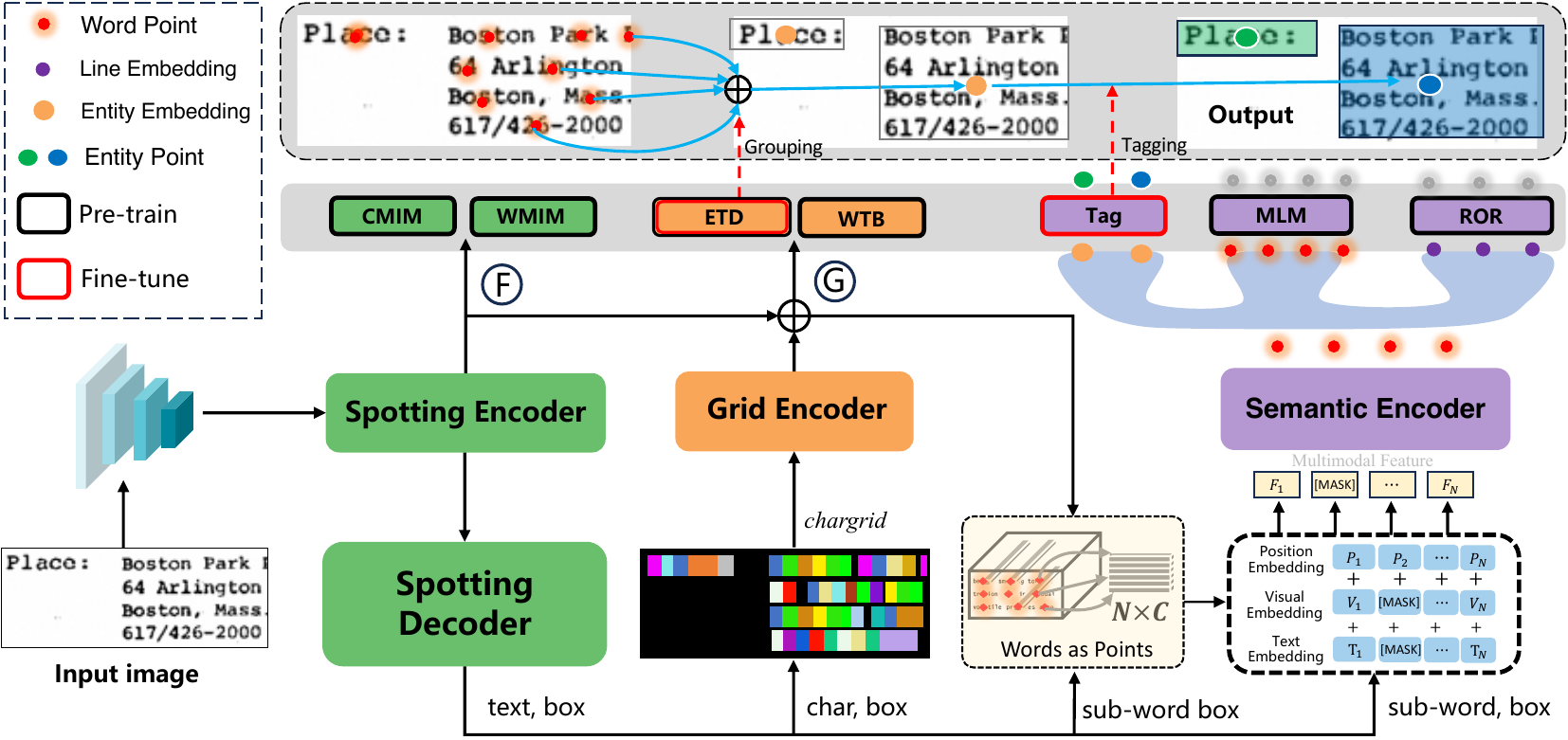}
  \caption{Schematic overview of HIP. The modules of the three hierarchical tasks are color-coded: green for spotting, orange for grouping, and purple for labeling. The top gray dashed box illustrates the main process from word point to entity point, where ETD and Tag branches play the roles of word grouping and entity labeling respectively.
  }
  \label{fig:fig2}
  
\end{figure*}

\vspace{-3mm}
\subsection{Overall Architecture}
As illustrated in Fig.\ref{fig:fig2}, the proposed \ourmodel is divided into three parts: word spotting (green), word grouping (orange), and entity labeling (purple). 

\textbf{End-To-End Framework.} In text spotting, after receiving image features from an initial pyramid CNN encoder, a text spotting encoder generates center proposals. Then, a spotting decoder produces the position and text content of a word by decoding the sequence of points from its center line. The encoding and decoding of center points are the same with DeepSolo~\cite{ye2023deepsolo} by simplifying curves into straight lines. In word grouping, the $chargrid$ is combined with feature $F$ after being encoded by the grid encoder to obtain feature $G$. Then, based on feature $G$, an entity detection module (ETD) is used to detect the bounding box of an entity. The $chargird$, which generates grid features, utilizes character-level text and position results from the spotting decoder (inference) or ground truth (training) in different stages. It ingeniously incorporates the textual information to enhance its representation. By extracting the center feature of each word in $G$ as visual embedding ($N$ × $C$) input for the entity labeling.

In entity labeling,  the features are first upgraded to multi-modal features by a semantic encoder, in which language is aligned with the text embedding from the ground truth (spotting decoder in inference) and visual embedding extracted from the feature $G$ by pre-training tasks. Finally, the entity tag head decodes these multi-modal features to produce the final entity categories. 

Therefore, word spotting obtains the location and text of words, and then groups words belonging to the same entity into an entity through word grouping. Finally, we obtain the category of the entity through entity labeling, thus achieving end-to-end information extraction.

\textbf{Pre-training Tasks.} The effectiveness of HIP, aside from the aforementioned end-to-end design, crucially stems from the representative encoders at each stage. We propose hierarchical pre-training strategies to encourage the model to understand complex visual, geometric, and semantic clues.  For the text spotting task,  we adopt image reconstruction as the pre-training task and introduce char-level and word-level mask image modeling (CMIM and WMIM) tasks to encode text features from various granularity. For the word grouping task, we devise top-down task entity detection (ETD) and bottom-up task word-to-block (WTB) to facilitate the learning of layout information. For entity labeling,  mask language modeling (MLM) and reading order reasoning (ROR) tasks are employed to strengthen the model's linguistic capabilities. 
% For word spotting, we adopt the char-level mask image modeling(\textbf{CMIM}) and word-level mask image modeling(\textbf{WMIM}). CMIM and WMIM primarily affect the spotting encoder. For word grouping, we propose an entity detection(\textbf{ETD}) task to capture the relationship from entities to words, and a word-to-block(\textbf{WTB}) task is proposed to model the relationship from words to entities. For entity labeling, we convert 3D grid features into 2D embeddings and feed them into the language model (semantic encoder). The semantic encoder is pre-trained by mask language modeling(\textbf{MLM}) and reading order reasoning(\textbf{ROR}). 
%The pre-training tasks will be detailed in the following sections.
\vspace{-3mm}
\subsection{Mask Image Modeling for Word Spotting}
%Although Masked Autoencoder(MAE)~\cite{he2022masked} is proven to be effective in general scenarios, they face two issues in OCR and VIE, due to the following reasons caused by the small size of text: 1) Randomly masking patches may result in large portions of text being masked, preventing the model from learning textual information. 2) The influence of the background may distract the model from fully focusing on learning about the text. To address the second issue, we adopt a strategy of masking only the text regions. For the first issue, we solve through a controllable hierarchical masking strategy, including CMIM and WMIM.

Unlike MAE~\cite{he2022masked}, there are three challenges of image-level document reconstruction: (1) the imbalanced foreground-background distribution,
(2) the difficulty of reconstruction varies largely between foreground and background, and (3) the reconstruction of characters and words also involves learning semantic representations.
Therefore, we only randomly mask and reconstruct the character and word regions in the images, which are denoted as CMIM and WMIM. 
%Although Masked Autoencoder~\cite{he2022masked} has been proven effective in general scenarios. Due to the imbalanced distribution of foreground and background, performing reconstruction in document-image presents some challenging issues: 1) randomly masking patches may lead to large portions of text being masked, which prevents the model from learning textual information; 2) the background may distract the model from fully focusing on learning about the text. To address those issues, we only randomly mask and reconstruct the character and word regions in the images, which are denoted as CMIM and WMIM. 

The CMIM randomly masks characters within a word. The CMIM guides the model to understand which characters constitute a specific word and what the corresponding character shapes are. %, which may correct some characters that are hard to recognize or prone to misrecognition, ultimately enhancing the accuracy of character recognition. 
The WMIM randomly masks words within a sentence. The WMIM encourages the model to observe the context and understand the textual semantics when asked to recover the masked words. Compared to CMIM, WMIM focuses more on contextual semantics, which involves learning some basic knowledge such as grammar and phrases. It is meaningful for spotting tasks, grouping, and labeling tasks, where an understanding of the text is required.

For the MIM head, we adopt a 2-layer transformer encoder with 8-head self-attention, 256 hidden size, and 1024 intermediate size of MLP to encode feature $F$, and then the encoded features are deconvolved to the original image size. The WMIM and CMIM share parameters completely. The masking ratio is set to 15\% and 30\% respectively, with an observation that WMIM is much more challenging to learn. The MIM is optimized with the cross-entropy loss function between the original image and the reconstructed one. The loss ${\mathcal{L}_{\text{MIM}}}$ is a linear combination of the loss of CMIM and WMIM: 
\begin{equation}
  \mathcal{L}_{\text{MIM}} = \lambda_{\text{CM}} * \mathcal{L}_{\text{CMIM}} + \lambda_{\text{WM}} *  \mathcal{L}_{\text{WMIM}},
\end{equation}
\noindent where $\lambda_{\text{WM}}$ and $\lambda_{\text{CM}}$ are coefficients and are set to 0.35 and 0.5 empirically.

\subsection{Entity Detection for Word Grouping}

We utilize a detection-based method to localize the boundaries of the entity. To obtain precise boundaries, we employed two strategies: 1) Casting text with spatial information into a $chargrid$ to capture rich linguistic information for entity detection; 2) Combining two pre-training tasks, bottom-up (Word to Block, WTB) and top-down (Entity Detection, ETD), to enable the model to fully learn layout information in the pre-training dataset.

% \vspace{0.5mm}
\textbf{Grid Encoder} is used to encode the character grid for entity detection. $D = \{(t_{k}, b_{k}|k=0,...,n\}$ represents texts, ${t_{k}}$ denotes the ${k}$-th character and ${b_{k}}$ is the corresponding box. The grid $g \in \mathbb{R}^{H \times W \times 3}$ is defined as
\begin{equation}
  g_{\text{i,j}} = \left\{
  \begin{aligned}
  E(t_{k}) , \quad if \;  (i,j) \;  \prec \;  b_{k} \\
  E([PAD]) , \quad   \text{otherwise}
  \end{aligned},
  \right.
  \label{eq:grid}
\end{equation}
where $\prec$ means coordinates [i, j] in the bounding box of $b_{k}$. $E$ represents the embedding operation, which maps character into feature space. $H$ and $W$ denote the height and width of the input image. The background pixels of the image are set as the embedding of a special token [PAD]. The final size of $chargrid$ is identical to the original image. Consequently, the $chargrid$ is processed as an image and a ViT \cite{dosovitskiy2020image} is used as the encoder to extract its features, denoted as Grid Encoder in Fig.\ref{fig:fig2}. Grid Encoder is composed of a 3-layer transformer encoder with 8-head self-attention, 256 hidden size, and a 1024 intermediate size of MLP. The patch size is set to 32. The features encoded by ViT are upsampled 4 times and then fused with the feature $F$ (the output of the FPN) and obtain $G \in \mathbb{R}^{h \times w \times 256}$,  $h = \frac{H}{8}$, and $w = \frac{W}{8}$ for entity detection. Note that, during training, $chargrid$ is generated with ground truth, while in inference, it is generated using the text contents and boxes predicted by the spotter.

% \vspace{0.5mm}
\textbf{ETD} is used for entity bounding box detection in both pre-training and fine-tuning stages. 
% During the pre-training stage, the boundaries of entities are pseudo, being delineated by the document's content and layout (such as colons and indentation), which forms the red boxes as shown in the ETD branch of Fig.~\ref{fig:fig1}. In the inference stage, these entity boundaries correspond to the exact boundaries of the entities.
During the pre-training stage, the pseudo boundaries of the entities are generated based on the document's content and layout (such as colons and indentation), resulting in the red boxes as shown in the ETD branch of Fig.~\ref{fig:fig1}. In the fine-tuning stage, the entity boundaries are defined by the ground truth.

Following the previous work CenterNet \cite{zhou2019objects}, there are three heads in ETD: the heatmap head, the size head, and the offset head. After upsampling $G$ by a factor of 2, it is fed into the three branches to get the entity's center point, width and height, and offset, respectively. With the three outputs, the entity bounding boxes can be computed easily. All branches are the same structure, comprising a 3x3 convolutional layer, a ReLU layer, and a 1x1 convolutional layer. Therefore, the total loss of the ETD task is formulated as
\begin{equation}
  \mathcal{L}_{\text{ETD}} = \mathcal{L}_{\text{Heatmap}} + \lambda_{\text{Size}}\mathcal{L}_{\text{Size}} + \mathcal{L}_{\text{Offset}},
\end{equation}

\noindent where $\lambda_{\text{Size}}=0.2$ in pre-training stage, and $\lambda_{\text{Size}}=1.0$ in fine-tuning stage.

% \vspace{0.5mm}
\textbf{WTB} defines the bottom-up criterion by learning whether a word belongs to an entity block. Compared to ETD, WTB exhibits improved performance in handling isolated words or words that span across lines.
%thereby forming a closed loop with the ETD task. 
%Although we have utilized the ETD task to achieve entity detection, defining the exact boundaries of an entity is still challenging.   
In this paper, WTB learns a relation matrix to determine whether a word belongs to an entity block. Firstly, two branches are designed to represent words and entities. Both branches have the same network structure, consisting of a 3x3 convolutional layer, a ReLU layer, and a 1x1 convolutional layer. The number of channels for the representations is 64. Taking $G$ as input, the word and entity are represented as $Rep_{word}$ and $Rep_{entity}$, with the dimensions of ${h \times w \times 64}$.

Assuming that there are $N$ words and $M$ entities in a given image, we extract word features from $Rep_{word}$ at the centers of words as their representations, obtaining ${W}^{N \times 64}$. The entity features are also extracted from $Rep_{entity}$ at the centers of entities, resulting in ${E}^{M \times 64}$. $W$ and $E$ are then reshaped into ${W}^{N \times 1 \times 64}$ and ${E}^{1 \times M \times 64}$, respectively. Finally, the $W - E$, in the shape of $(N \times M \times 64)$, is transformed to the output matrix as ${O}^{N \times M}$ with a 1x1 convolutional layer. The value in the i-th row and j-th column of matrix $O$ is supposed to be 1 if the $i_{th}$ word is inside the $j_{th}$ entity. We utilize the cross-entropy loss function for supervised learning, and the loss is denoted as $L_{\text{WTB}}$.

\subsection{Entity Labeling}
\textbf{Semantic Encoder.} 
%The MLM task guides the model to understand the context semantics by having it fill in the blanks. By implicitly learning the reading order of the text, the ROR task helps the model understand the text's layout and relative positional relationships. 
The Semantic Encoder consists of a 6-layer transformer with 4-head self-attention, 256 hidden size. The visual embeddings input to the Semantic Encoder are the word center features (with a size of $N \times C$) extracted from the Grid Encoder, as shown in the light yellow box in Fig.~\ref{fig:fig2}. The text embeddings are different in training and inference. In ference, texts are generated by the spotting decoder, while in training, they are ground truth. 
%The visual embeddings of sub-words belonging to the same word are represented by the visual embedding of the word. 
Following LayoutLMv3~\cite{huang2022layoutlmv3}, position embedding is added to the input of the Semantic Encoder. These embeddings are organized in a sequence from left to right and top to bottom in space. To enhance the linguistic representativeness of the semantic encoder in HIP, two pre-training tasks, MLM and ROR are adopted. 
% in spatial coordinates.

% \vspace{0.5mm}
\textbf{MLM} guides the model to understand the context semantics by recovering the masked tokens~\cite{xu2020layoutlm}. We randomly select 15\% of the words to implement MLM (i.e., the words masked in WMIM task in an image). Among them, 80\% are replaced with [MASK], 10\% are randomly replaced with other tokens, and 10\% remain unchanged. A fully connected layer is added to the Semantic Encoder to predict the masked tokens.

% \vspace{0.5mm}
\textbf{ROR} helps the model understand the text's layout and relative spatial relationships by implicitly learning the reading order of the text lines.
%\red{Clearly explain what capabilities the ROR module enhances. grouping alone is not sufficient, whether ROR helps in determining the correct order within a group. so does the ROR module helps forming correct order?}
%captures the relative positional relationships of fields within an image. 
Reading order is crucial for VIE, 
as values often appear to the right or below their corresponding keys. The prior knowledge benefits the entity labeling task.
In the ROR task, we focus on the relative positions of text lines. The % reading order in the paragraph level can be obtained in Algorithm 1, and the 
pseudo label of reading order at the line level can be obtained by sorting from top to bottom and left to right. Notably, the input embeddings of the Semantic Encoder are in the sub-word level,  while the ROR task involves reasoning at the line level.  The first word embedding of each line from the Semantic Encoder is adopted as the representation of the entire line. A visualization of ROR is depicted in the figure below the ROR module in  
Fig.~\ref{fig:fig1}. Each turning in the blue line represents a reading order change. 
% The paragraph information, sorted from top to bottom and left to right, has been obtained in \hyperref[alg:algorithm1]{Algorithm 1}. This sequence essentially represents the reading order of the paragraphs. By incorporating the order of lines within each paragraph, we can obtain the reading order at the line level. Notably, the input to the Semantic Encoder is only word-level embeddings sorted by spatial position, while the ROR task involves reasoning at the entity (paragraph) level reading order, the order of these two is not the same concept.

% Paragraph information has already been generated in ETD. We use two steps as follows to obtain pseudo labels for the ROR task:

% \vspace{-2.0mm}
% \begin{itemize}
%     \item[-] {Discretize the paragraphs into segments (Each line can also be referred to as a segment); }
%     \item[-] {Sort the paragraphs and then the segments within the paragraphs in order from top to bottom and left to right. }
% \end{itemize}
% \vspace{-2.0mm}

The global reading order of text lines is converted into local relations of whether the lines are adjacent. Assuming N lines are given in an image, and their corresponding features are obtained from the Semantic Encoder, with a size of $N \times 64$; Following the WTB  head, we propose a ROR head to construct an $N \times N $ matrix with a fully connected layer that represents the adjacency between any two lines. Only the current line and its next line are considered adjacent. The optimization goal of adjacent lines is set to 1, and the rest are set to 0. Here we use $L_{\text{MLM}}$ and $L_{\text{ROR}}$ to represent the losses for the MLM and ROR tasks, respectively.

% \vspace{-3.0mm}
% \subsection{Hierarchical Point Representation} 

% Besides decomposing the VIE task into hierarchical subtasks, we also construct our model using hierarchical representations of points. In the Text Spotting phase, we sample the top-K points from the central line to represent a word, which can efficiently encode the position, shape, and semantics of the text. These points are at the char-level. Subsequently, we extract the central points of all words as visual representations, efficiently converting 3D grid features into 2D embedding, thus smoothly integrating into the language model for contextual semantic understanding. Here, the points are at the word-level. Finally, using the entity boxes obtained from the ETD task, we average all word embeddings within an entity to represent corresponding entity. These points are at the entity-level.

% Therefore, from char-level to word-level to entity-level, we utilize points of three different levels to represent information for tasks at various levels, achieving an information representation that progressively aggregates from lower to higher levels. Hierarchical point representation not only simplifies the model but also enhances its interpretability.

\subsection{Training Objectives}
In this section, the pre-training loss $\mathcal{L}_{\text{PT}}$ and fine-tuning loss $\mathcal{L}_{\text{FT}}$ will be detailed. For the word spotting task, we employed the loss from DeepSolo~\cite{ye2023deepsolo}, denoted as $L_{\text{Spot}}$. Ultimately, the total loss for the pre-training stage is:

\begin{equation}
  \mathcal{L}_{\text{PT}} = \mathcal{L}_{\text{Spot}} + \mathcal{L}_{\text{MIM}} + \mathcal{L}_{\text{ETD}} + \mathcal{L}_{\text{WTB}} + \mathcal{L}_{\text{MLM}} + \mathcal{L}_{\text{ROR}},
\end{equation}

\noindent In the fine-tuning stage,
the final result of entities consists of three parts: entities' locations, entities' contents, and entities' categories. Entities' locations are generated by ETD. Entities' contents are the words within an entity. Entities' categories are generated by the Tag head and the loss is denoted as $\mathcal{L}_{\text{Tag}}$.
We only train the word spotting, word grouping, and entity labeling tasks in the fine-tuning stage,
%which are defined as $\mathcal{L}_{\text{Spot}}$,  $\mathcal{L}_{\text{ETD}}$, and $\mathcal{L}_{\text{Tag}}$ above.  
% Entities are formed by combining words output by Spotting within the bounding boxes obtained from ETD, which are then input into the Tag head to obtain entity categories, thereby achieving information extraction. 
%To enrich the model's learning, we supervise not only the classification of entire entities with $L_{\text{B}}$ but also assign the category information of the respective entities to each word and supervise their learning with $L_{\text{W}}$. During inference, only block-level results are utilized. 
and the loss function $\mathcal{L}_{\text{FT}} $ is formulated as

\begin{equation}
  \mathcal{L}_{\text{FT}} = \mathcal{L}_{\text{Spot}} +  \mathcal{L}_{\text{ETD}} +  
 \lambda_{\text{Tag}} * \mathcal{L}_{\text{Tag}},  
  % \lambda_{\text{B}}\mathcal{L}_{\text{B}} + \lambda_{\text{W}}\mathcal{L}_{\text{W}} + 
  % \mathcal{L}_{\text{ETD}} + , 
\end{equation}

\noindent where $\lambda_{\text{Tag}}$ is set to 10, empirically. 

% Additionally, we found that during the fine-tuning stage, omitting text embedding and using only the central point features of each word as visual embedding, combined with position embedding as inputs to the Semantic Encoder, and employing the randomly initialized parameters of the Semantic Encoder yields the best results.

\begin{table*}[tb]
\centering
\begin{tabular}{lcccccc}
\toprule
\multicolumn{1}{l}{\multirow{2}{*}{Methods}} & \multicolumn{1}{l}{\multirow{2}{*}{\#Param.}} & \multicolumn{1}{c}{FUNSD} & \multicolumn{2}{c}{CORD} & \multicolumn{2}{c}{SROIE}  \\
\cmidrule(lr){3-3} \cmidrule(lr){4-5} \cmidrule(lr){6-7}
& &F1-score  & F1-score & Accuracy &  F1-score & Accuracy  \\
\midrule 
TRIE~\cite{zhang2020trie}      & -       & -  & -    & -     & 82.1  & -   \\
Donut~\cite{kim2022ocr}      & 143M         &${28.0^{\star}}$   & 84.1 & 90.9   & 83.2  &92.8 \\
ESP~\cite{yang2023modeling}   & 50M        &${34.0^{\dagger}}$  & - &- &-   &- \\
Dessurt~\cite{davis2022end}   & 87M        & -   & 82.5 & -     & 84.9  & -   \\
DocParser (Dhouib et al.2023) & 70M  & -   & 84.5 & -     & 87.3  & -   \\
SeRum~\cite{cao2023attention}   & 136M     & -   & 80.5 & 85.8  & 85.6  &92.8 \\
% \textcolor{gray}{SeRum\textsubscript{prompt}~\cite{cao2023attention}*}   & No  & \textcolor{gray}{84.9} & \textcolor{gray}{91.5}  & \textcolor{gray}{85.8} & \textcolor{gray}{95.4} \\
% CFAM~\cite{kuang2023visual}          & $-{\text{M}}$   & -    & -    & - & -    & 85.87 & -    \\
\midrule
HIP(ours)         & 112M       &\textbf{52.0}~(\textcolor{red}{+18})   &\textbf{85.7}~(\textcolor{red}{+1.2}) & \underline{86.5} & \textbf{87.6}~(\textcolor{red}{+0.3}) & \textbf{96.7}~(\textcolor{red}{+3.9}) \\

\midrule
\textcolor{lightgray}{Qwen-VL-Max (qwen-vl-max-0809)} &\textcolor{lightgray}{-}  & \textcolor{lightgray}{19.0}   & \textcolor{lightgray}{45.0}  & \textcolor{lightgray}{71.5} & \textcolor{lightgray}{66.1} & \textcolor{lightgray}{82.0}   \\
\textcolor{lightgray}{GPT-4V (gpt-4-turbo-2024-04-09)}  &\textcolor{lightgray}{-}  & \textcolor{lightgray}{45.5}  & \textcolor{lightgray}{68.1}  & \textcolor{lightgray}{87.7} & \textcolor{lightgray}{70.4} & \textcolor{lightgray}{92.2}  \\
\textcolor{lightgray}{GPT-4o (gpt-4o-2024-05-13)}  & \textcolor{lightgray}{-} & \textcolor{lightgray}{47.2} & \textcolor{lightgray}{82.4} & \textcolor{lightgray}{94.2} & \textcolor{lightgray}{77.2} & \textcolor{lightgray}{92.2}      \\
\bottomrule
\end{tabular}
\caption{Comparisons of end-to-end methods with OCR-free metrics on FUNSD, CORD, and SROIE. The ${\star}$ denotes the results obtained by training and testing with the official code. The ${\dagger}$ indicates that the result is reproduced with word spotting results and word grouping boxes from the proposed HIP. The results of closed-source large models are set to light gray. 
}
\label{table_kie}
\end{table*}

\begin{table}[tb]
\small
\centering
\begin{tabular}{lcccc} 
\toprule
Methods  &Spotting  &Grouping   &Labeling  & EE \\
\midrule
$\rm StrucTexT_{Base}^{\star}$    &-  &-  &-    &46.8 \\
$\rm LayoutLMv3_{Base}^{\star}$ &-  &-  &-   &53.1  \\
$\rm StrucTexTv2_{Small}^{\star}$  &-  &-  &-    &{55.0} \\
$\rm StrucTexTv2_{Small}^{\dagger} $  &80.0   &81.4   &87.3     &{55.1} \\
\midrule
$\rm HIP~(ours)$  &\textbf{81.3}  &\textbf{84.8}   &\textbf{88.1}                 &\textbf{60.3} \\
\bottomrule
\end{tabular}
\caption{Comparisons on FUNSD with OCR-based metrics. The ${\star}$ denotes the results are reported by StrucTexTv2, while ${\dagger}$ means the results are measured with the official open-sourced model. The EE is short for Entity Extraction.
% Language-based methods are only evaluated on entity tagging. EL denotes entity linking, and ET denotes entity tagging.
%`Accuracy' denotes the tree-edit-distance-based accuracy. 
}
\label{table:funsd}
\end{table}

\begin{table}[tb]
\centering
\begin{tabular}{lcccc} 
    \caption{Ablation study for the effectiveness of 
hierarchical pre-training tasks. S, G, L, and EE are the abbreviations of Spotting, Grouping, Labeling, and Entity Extraction in order. The OCR-based metrics are adopted.}
    \label{tab:ab_pretrain}
    \scalebox{0.65}{
    \begin{tabular}{|c|c|c|c|c|c|c|c|c|c|}
    \hline
        No. & MIM & ETD & WTB & ROR & MLM & S & G & L &EE\\
        \hline
        \textbf{(a)} & \checkmark & \checkmark & \checkmark & \checkmark & \checkmark & 81.3  & 84.8  & 88.1  & 60.3~(\textcolor{red}{+9.7}) \\ 
        \textbf{(b)} & \checkmark & ~ & ~ & ~ & ~ & 82.1  & 81.4  & 86.1  & 54.2~(\textcolor{red}{+3.6}) \\ 
        \textbf{(c)} & ~ & \checkmark & ~ & ~ & ~ & 81.5  & 85.3  & 84.8  & 59.0~(\textcolor{red}{+8.4}) \\ 
        \textbf{(d)} & ~ & ~ & \checkmark & ~ & ~ & 79.9  & 84.8  & 82.1  & 56.2~(\textcolor{red}{+5.5}) \\ 
        \textbf{(e)} & ~ & ~ & ~ & \checkmark & ~ & 80.2  & 80.3  & 88.3  & 54.2~(\textcolor{red}{+3.5}) \\ 
        \textbf{(f)} & ~ & ~ & ~ & ~ & \checkmark & 82.6  & 81.2  & 87.0  & 55.5~(\textcolor{red}{+4.9}) \\ 
        \textbf{(g)} & ~ & ~ & ~ & ~ & ~ & 79.1  & 81.8  & 80.5  & 50.6 \\ \hline
    \end{tabular}
    }
\end{tabular}
\end{table}

\section{Experiments}
\subsection{Datasets and Metrics}

% Other works\cite{kim2022ocr, cao2023attention, davis2022end}, in addition to using the IIT-CDIP~\cite{lewis2006building}, employ a variety of other data sources; for instance, \cite{kim2022ocr} utilized their own tool, SynthDoG, to generate 2 million samples for pre-training. Consequently, besides IIT-CDIP, HierText~\cite{long2022hiertext} is used for our pre-training.
The IIT-CDIP~\cite{lewis2006building} and HierText~\cite{long2022hiertext} are used for pre-training. 
CORD~\cite{park2019cord}, FUNSD~\cite{jaume2019funsd}, and SROIE~\cite{huang2019icdar2019} are adopted as benchmarks
% to compare with state-of-the-art methods 
to validate the effectiveness of HIP on VIE.

\textbf{Pre-Training.} The IIT-CDIP Test Collection 1.0 dataset contains 11 million multi-page documents, with a total of 42 million pages. In the pre-training phase, we removed rotated images and used 6.5 million images for training. HierText consists of 8281 training samples and 1724 testing samples and contains high-quality word, line, and paragraph-level annotations.
%featuring hierarchical annotations of text in natural scenes and documents. The dataset 
%, where images average more than 100 words per image. 
Only the training split is employed for training.

\textbf{Fine-Tuning.} Three benchmark datasets are adopted to evaluate our HIP. The \textbf{FUNSD} is composed of 149 training samples and 50 testing samples, where the images are scanned English documents with four predefined entities, namely “key”, “value”, “header”, and “other”.
The \textbf{CORD} consists of 30 labels across 4 categories. The train, validation, and test splits contain 800, 100, and 100 samples, respectively. The \textbf{SROIE} dataset comprises a training set with 626 receipts and a testing set with 347 receipts. 
%Each receipt in the dataset contains four predefined entities, namely: “company”, “date”, “address”, and “total”. 
To meet the word-level annotation granularity required by the model,  we use spaces to segment the transcriptions and manually modify the line-level annotation into word-level annotation. 

% The annotations in the dataset provide entity-level and segment-level bounding boxes for the text regions and their corresponding transcriptions. To meet the word-level annotation granularity required by the model,  we use spaces to segment the transcriptions and manually modify the line-level annotation into word-level annotation. 

\textbf{Metrics.} Currently, there are two kinds of methods for evaluating VIE, namely OCR-based and OCR-free metrics.
% , which are often used in OCR-based and OCR-free methods
The difference lies in whether the detection part is considered. For OCR-based metrics, the entity extraction task comprises word spotting, word grouping, and entity labeling. Following ~\cite{yu2023structextv2}, we adopt the Normalized Edit Distance(1-NED) for evaluating word spotting and entity extraction. Word grouping is formulated as a one-class detection evaluated with F1-score, and entity labeling is evaluated with Accuracy as a multi-classification. For OCR-free metrics, we evaluate the models with entity-level F1-score and Tree Edit Distance (TED) based accuracy, consistent with ~\cite{kim2022ocr}, abbreviated as F1-score and Accuracy in this paper.

\subsection{Implementation Details}
% Due to the excellent performance on text spotting,
The details of the network structure can be referred to the architecture section. Data augmentation employs operations such as cropping, scaling, and noise rendering, which are consistent with the practices adopted in text spotting.
% The number of heads and sampling points for deformable attention is 8 and 4, respectively. The number of both encoder and decoder layers is 6. The number of sampled points N is set to 25 by default, and the model predicts 96 character classes without special instructions. 
% The word features, which belong to the same group and come out of the semantic encoder, are averaged to obtain the grouping representation. An additional 2-layer MLP block is performed with the grouping representation to classify entity labels. 

\textbf{Pre-Training.} We train our models with 6.6 million samples with all pre-training tasks activated. The learning rate is set to 1e-4 without decay. The number of proposals $K$ in the spotting encoder is set to 100. For data augmentation, the longest side of the training image does not exceed 1024, and the shortest side is randomly selected from [768, 800, 832, 864, 896]. With the batch size set to 96 in 8 Nvidia Tesla 80G A100 GPUs, the pre-trained model is trained for 285,000 steps and is used as parameter initialization in the fine-tuning task.

\textbf{Fine-Tuning.} The fine-tuning tasks are performed on each evaluation data set separately. The number of proposals $K$ in the spotting encoder is set to 150 for CORD and SROIE, and 512 for FUNSD. For data augmentation, the longest side of the image does not exceed 2500, and the shortest side is randomly selected from [1000, 1200, 1400, 1600]. The fine-tuning tasks are trained for a total of 120,000 steps, with the learning rate decaying at the 80,000th step. 
%The fine-tuning tasks are trained for a total of 120,000 steps in a single Nvidia Tesla 80G A100 GPU with batch size set to 2. The learning rate is initialized to 1e-4, and decayed by a factor of 0.1 at the 80,000th step. 
For evaluation, the shortest and longest sides are set to 1600 and 2900. The confidence threshold for word spotting is set to 0.3, and the IOU threshold for word grouping is set to 0.4, empirically.

\subsection{Comparison with the State-of-the-Arts}
% In this section, we evaluate the entity extraction performance of HIP with several state-of-the-art techniques on different benchmarks. 
% As mentioned in ~\hyperref[paragraph:Metrics]{Metrics}, different metric styles are adopted in different papers. 
For a comprehensive comparison, we use both OCR-based and OCR-free metrics for the FUNSD. However, due to the lack of results from previous works, the CORD and SROIE benchmarks are only evaluated with OCR-free metrics.

\textbf{Comparisons on Benchmarks with OCR-free Metrics.} We compare HIP with OCR-based methods~\cite{zhang2020trie,yang2023modeling} and OCR-free methods~\cite{kim2022ocr,davis2022end,dhouib2023docparser,cao2023attention} on three public benchmarks in OCR-free metrics, and the results are summarized in Tab.\ref{table_kie}. The proposed HIP achieves the SOTA on almost all public benchmarks. In terms of F1-score, HIP outperforms the previous SOTA methods by 18\%, 1.2\%, and 0.3\%. In terms of Accuracy, HIP surpasses Donut and SeRum on SROIE with an increase of 3.9\%. HIP ranks second on the CORD because some of the key-value structures are too complex to parse well. 
% which proves once again the leadership of our method. The FUNSD is more challenging due to its longer and multi-line entities, and the encoder-decoder methods lack an advantage in decoding long fields, resulting in few reports of results.
It is worth noting that the Accuracy on FUNSD is not reported because the entity categories are defined as “question”, “answer” etc., which are meaningless to be represented as tree structures. We also evaluate the performance of large language models and observe that only GPT-4O demonstrates superior accuracy on CORD. We attribute this improvement to its enhanced linguistic capabilities, enabling it to parse more complex key-value structures.
%The CORD dataset has a rich hierarchical structure, which we have not explored in depth, resulting in a second-best accuracy for our method on this dataset.

\textbf{Comparisons on FUNSD with OCR-based Metrics.} We compare HIP with $\rm LayoutLMv3_{Base}$~\cite{huang2022layoutlmv3} and StrucTexT~\cite{li2021structext, yu2023structextv2} on FUNSD benchmark with OCR-based metrics, and the results are summarized in Tab.~\ref{table:funsd}. We observe that HIP outperforms all the previous methods in the final entity extraction metric. HIP boosts the upper bound of FUNSD to a new record of 60.3\%, demonstrating its effectiveness. To analyze the performance of each subtask, we modify the official code released by StrucTexTv2~\cite{yu2023structextv2} to support the evaluation of word spotting, word grouping, and entity labeling. HIP outperforms $\rm StrucTexTv2_{Small}^{\dagger}$ by 1.3\%, 3.4\%, and 0.8\% on the three sub-tasks. The original report of StrucTexTv2 only reports the performance of the small version, so we do not compare it with a larger version.

\subsection{Ablation Study}
% We decompose the entity extraction task into three subtasks: word spotting, word grouping, and entity labeling, and propose three groups of hierarchical pre-training tasks: MIM, ETD \& WTB, and ROR \& MLM to improve the performance of each subtask, respectively. 
In this section, several experiments are conducted on FUNSD to verify the effectiveness of the hierarchical pre-training strategy, where the model is first pre-trained with various pre-training tasks, and then fine-tuned on FUNSD.
 % In the pre-training stage, we merged the MIM tasks of different granularities, CMIM and WMIM tasks, into one pre-training task to simplify the experiment,
In the pre-training stage, we merge the CMIM and WMIM tasks into one to simplify the experiment, resulting in a total of 5 pre-training tasks, as presented in Tab.~\ref{tab:ab_pretrain}. In experiments (a)-(g), the CNN weights are initialized using ImageNet~\cite{deng2009imagenet} pre-trained weights, and the OCR task is trained concurrently to ensure the updates to the spotting encoder.

 \textbf{Impact of Sub-tasks}. When studying the effects of experiments (a) and (g) across different metrics (columns S, G, and L), it is evident that the spotting is the lowest, with values of 79.1\% and 81.3\% respectively. This suggests that text spotting remains the primary bottleneck of the VIE task.
 % We use four Nvidia Tesla 80G A100 GPUs for training, the batch size is half to 48, and the other training configuration is consistent with the main pre-training task. 
% To efficiently verify the effectiveness of hierarchical pre-training tasks, we divide all pre-training tasks into three groups, denoted as spotting, grouping, and labeling, respectively. The spotting set includes CMIM and WMIM tasks, the grouping set includes ETD and WTB tasks, and the labeling set includes MLM and ROR tasks. As indicated in Tab.~\ref{tab:ab_pretrain}, the experiments (a)-(e) are trained with specific pre-training tasks with only the parameters of the backbone network initialized using ImageNet~\cite{deng2009imagenet} weights. We use four Nvidia Tesla 80G A100 GPUs for training, the batch size is half to 48, and the other training configuration is consistent with the main pre-training task. Due to GPU resource limitations, each pre-training task is only trained for 150,000 steps.
% %which takes approximately 15 days. 
%We finetune our model on FUNSD with the pre-training weights as initialization.

\textbf{Effects on Pre-training tasks}.
By comparing (g) and the others on EE, it can be observed that all pre-training tasks bring a total improvement of 9.7\%,  with ETD contributing the largest increase of 8.4\%. The ETD task also shows the greatest improvement in grouping metrics (column G), indicating that entity boundaries are crucial for end-to-end VIE. 

Next, we examine the effects of the pre-training tasks corresponding to different subtasks. (1) By comparing (b) and (g), it can be observed that the MIM task, designed specifically for text spotting, achieves a 3\% improvement in the spotting metric (column S). It is noteworthy that the MLM task contributes more significantly to the spotting metric. This is because the MLM task is closely aligned with the task of recognizing text within masked regions. (2) The effectiveness of ETD and WTB is verified by experiments (c) and (d), which bring the top 2 largest improvements on grouping (column G) by 3.5\% and 3.0\%, respectively. (3) By comparing experiments (e, f) with (g), it can be seen that pre-training ROR and MLM related to the entity labeling (column G) leads to the most significant improvements in the entity labeling metric, with increases of 7.8\% and 6.5\%, respectively. %The benefits of the ETD task, consistent with downstream tasks, are most significant, which indicates that using pseudo-labels to extract layout information from pre-training data is very meaningful for distinguishing entity boundaries. For the WTB task, the subordinate relationship from word to block is learned to implicitly improve the representation for word grouping.

\begin{figure}[t]
  \centering
  \includegraphics[height=3.5cm]{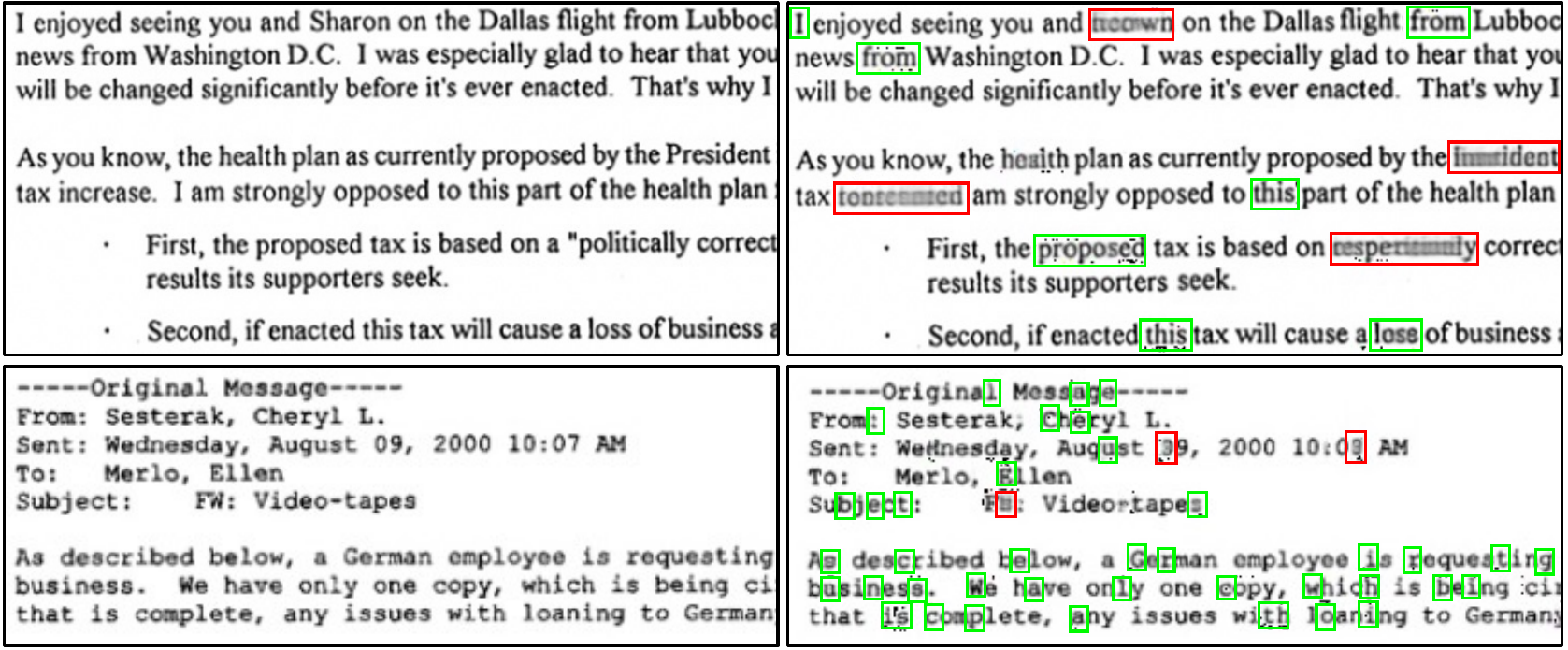}
  \caption{
  %The visualization of CMIM and WMIM. The first column shows the input image, and the second column displays the corresponding reconstruction image. The first row depicts the WMIM visualization, while the second row illustrates the CMIM visualization. The red box indicates a good case, and the blue box signifies a bad case.
  The visualization of MIM tasks. The first and second rows are the results of WMIM and CMIM, and the first and second columns are the original images and the reconstructed images respectively. The green box indicates a good case, and the red box signifies a blurred case.
  }
  \label{fig:fig3}
\end{figure}

% \begin{figure*}[t]
%   \centering
%   \includegraphics[width=0.7\textwidth]{figure/vis_qres_v5.pdf}
%   \caption{Qualitative results of our method on the FUNSD dataset. The columns from left to right represent the visualization of ground truth, StrucTexTv2, and HIP. The rectangles in green and red stand for correct and incorrect results, respectively. Dashed boxes indicate detection errors, while underlined characters signify the loss of recognition in word spotting.
%   }
%   \label{fig:vis_qres}
% \end{figure*}

\begin{figure}[t]
  \centering
  \includegraphics[width=1.0\textwidth]{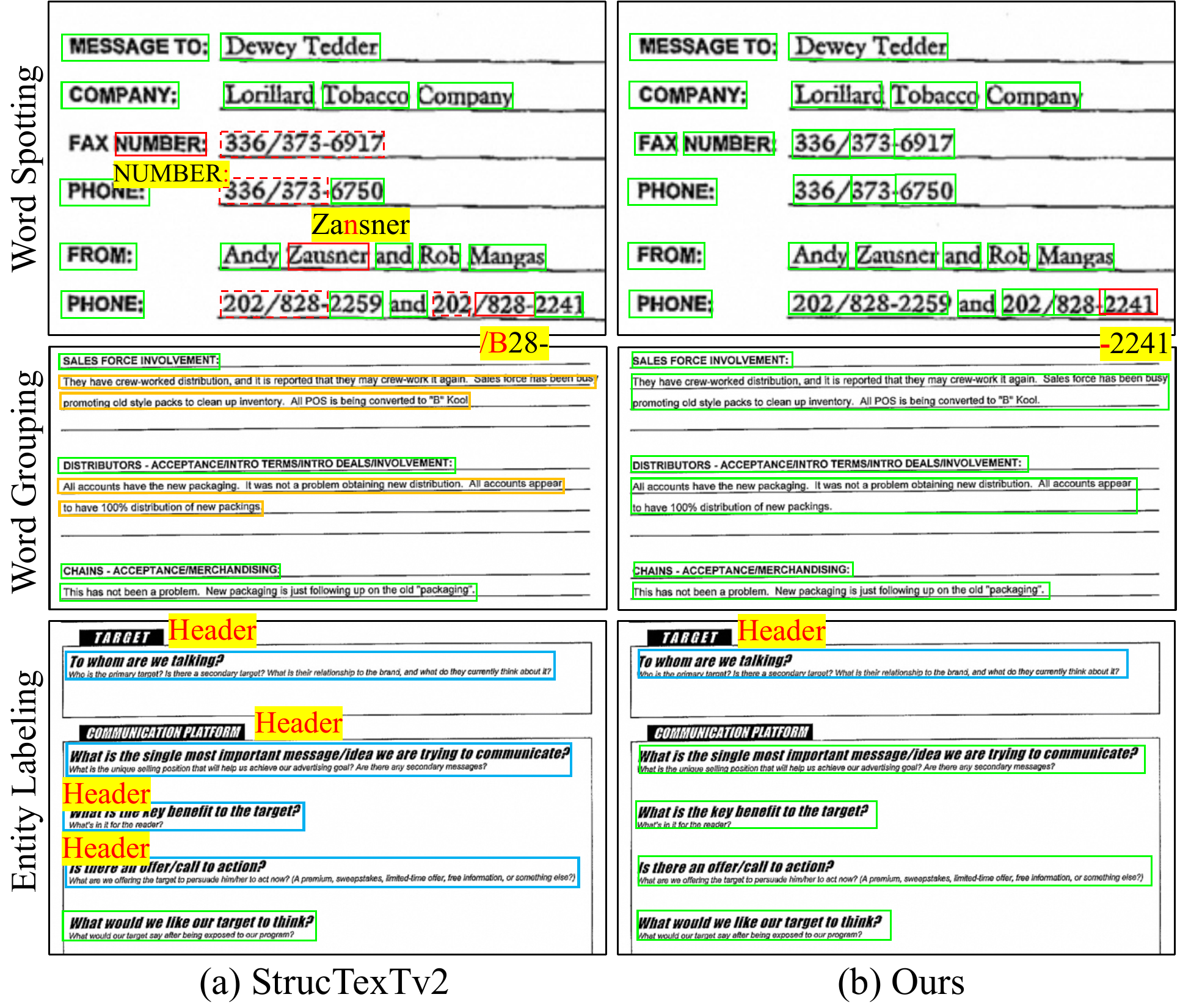}
  \caption{Qualitative results on the FUNSD. The columns from left to right represent the visualization of StrucTexTv2 and HIP. The rectangles in green stand for correct results.  The red boxes denote word spotting errors, where detection errors and recognition errors are represented by dashed and solid boxes, respectively. The yellow boxes denote word grouping errors. The blue boxes denote entity labeling errors and the misclassified categories are noted with red words.}
  % red characters denote recognition errors in word spotting.}
  \label{fig:vis_qres}
\end{figure}

\subsection{Qualitative Analysis}
\textbf{Visualization of MIM Tasks.} The reconstruction results for CMIM and WMIM are presented in Fig.~\ref{fig:fig3}. For the CMIM task, as shown in the second row, there are only three characters not successfully reconstructed, one of which is an abbreviation of an uncommon word, and the other two are numerals, which makes the failures reasonable. The reconstruction of characters shows that the model can capture the shape of characters and find characters to compose a word. Regarding the WMIM task, the frequently occurring short prepositions, conjunctions, and other words such as \emph{``from'', ``and'', ``am'', ``this''}, and \emph{``I''} are well reconstructed. This suggests the model has grasped basic grammar and understood the context. However, the reconstruction of proper nouns, adjectives, and verbs (\emph{``Sharon'', ``President'', ``increase''}, and \emph{``politically''}) is less effective. It can be attributed to the fact that these words often have numerous synonyms which do not significantly impact the sentence's grammaticality.
%Surprisingly, longer words such as \emph{``cigarette''} and \emph{``proposed''} are successfully reconstructed, likely because the model captured information related to \emph{``cigarette''} within the text, allowing for accurate reconstruction. 
% The ability to achieve such understanding from images alone benefits spotting, grouping, and labeling tasks, as evidenced by the comparison between experiments (b) and (f) in Tab.~\ref{tab:ab_pretrain}.

\textbf{Qualitative Results of the HIP on the FUNSD Dataset.} We perform a qualitative analysis of our HIP and StrucTexTv2 on FUNSD to show the superiority of our method. As presented in Fig.~\ref{fig:vis_qres}, the rows from top to bottom represent word spotting, word grouping, and entity labeling, respectively. The columns from left to right represent the visualization of StrucTexTv2 and HIP. In word spotting, StrucTexTv2 is more likely to be confused by characters with similar shapes; for example, the phone number \emph{“zausner”} is incorrectly recognized as \emph{“zansner”}. In word grouping, texts belonging to the same entity but located in different lines are prone to be recognized as two separate entities. In entity labeling, the category of "Question-answer" tends to have a similar appearance to the "Header", both often presented in bold text. StrucTexTv2 frequently misclassifies those entities, while HIP demonstrates a higher accuracy in assigning the correct labels. 

\section{Conclusion}
In this paper, we have proposed an end-to-end visual information extraction framework \ourmodel, which integrates word spotting, word grouping, and entity labeling. Taking the merit of point representations, \ourmodel models entities as hierarchical points to better fit the hierarchical nature of the VIE task. We have also introduced three kinds of pre-training strategies, which enhanced the model's representation capabilities and improved the final results. Experiments on standard benchmarks demonstrate that \ourmodel achieves highly competitive results, particularly excelling in stricter F-score metrics, indicating precise recognition accuracy. However, the study reveals the limitations of current VIE methods in handling low-quality and complex documents, highlighting a need for further research in these challenging scenarios.
%Inspired by point-based methods such as ESP and DeepSolo, we introduce a hierarchical point representation method to flexibly unify various granular tasks of information extraction within a single framework, thus proposing our HIP. Base on the nature that VIE task is inherently hierarchical, including Text Spotting, Word Grouping, and Entity Labeling, we propose hierarchical pre-training tasks to enhance the performance at each stage. Experiments demonstrate that HIP sets new state-of-the-art benchmarks across multiple datasets. The visualization of WMIM task shows that the model has learned basic grammar and phrases and can comprehend contextual semantics to accurately reconstruct masked words. 

% \clearpage\mbox{}Page \thepage\ of the manuscript.
% \clearpage\mbox{}Page \thepage\ of the manuscript.
% \clearpage\mbox{}Page \thepage\ of the manuscript.
% \clearpage\mbox{}Page \thepage\ of the manuscript.
% \clearpage\mbox{}Page \thepage\ of the manuscript. This is the last page.
% Now we have reached the maximum length of an ECCV \ECCVyear{} submission (excluding references).
% References should start immediately after the main text, but can continue past p.\ 14 if needed.
\clearpage  % TODO REVIEW/FINAL: This \clearpage needs to be removed from both review and camera-ready versions.

\bibliography{main}
\end{document}